\documentclass[11pt]{article}
\usepackage{acl2016}
\usepackage{times}
\usepackage{url}
\usepackage{latexsym}

\aclfinalcopy 

\title{Desiderata for Vector-Space Word Representations}

\author{Leon Derczynski \\
  Department of Computer Science \\
  University of Sheffield \\
  S1 4DP, UK \\
  {\tt leon.d@shef.ac.uk} \\}

\date{}

\begin{document}
\maketitle
\begin{abstract}
A plethora of vector-space representations for words is currently available, which is growing.
These consist of fixed-length vectors containing real values, which represent a word.
The result is a representation upon which the power of many conventional information processing and data mining techniques can be brought to bear, as long as the representations are designed with some forethought and fit certain constraints.
This paper details desiderata for the design of vector space representations of words.
\end{abstract}

\section{Introduction}

The following desiderata describe attributes that are useful to have in a vector space word representation, either because they provide a more compact or elegant way of presenting the data, or because they make other tasks easier.

\section{Geometry}
High intrinsic quality of the geometry of vector space (hereafter VS) representations is important.
A primary feature of VS representations support for algebraic reasoning over semantic concepts; e.g, the \emph{king-man+woman=queen}-style analogy results~\cite{mikolov2013distributed}.
Following~\newcite{schnabel2015evaluation} other intrinsic qualities include relatedness, analogy, categorisation and selection preference powers.
A good VS representation has a geometry that performs well in these criteria.

\section{Deterministic}
Multiple runs over same dataset produce the same representations.
That is to say, it should be deterministic, allowing for repeatability.
As there are no constraint that make randomness particularly beneficial, introducing it into representation construction can lead to unnecessary destabilisation of the results.
For example, should be no need to hill-climb to find the best representation, which might otherwise benefit from starting from many different points distributed randomly.
Rather, the same representation should always emerge from the same data and hyperparameters regardless of internal starting states.
Further, decisions made during representation generation should always play out the same way.
A wider range of analysis methods can be brought to bear if VS representations are deterministic.

\section{Fully-specified}
Following from the point on determinism, algorithms should be specific as fully as possible, in order to ensure accurate reproduction.
The best specification is both an abstract explanation of the approach taken and also matching source code (sometimes a problem, e.g. with the original Brown clustering description).
This includes handling things such as tie-breaking, which are not always addressed well, e.g. in Brown clustering~\cite{derczynski2016generalised}.
In any event, an algorithm for vector space generation should have no unspecified edge case handling, including e.g. ambiguous tie-breaking, or requiring a random seed.

\section{Semantic closeness}
Just as a high quality VS geometry has good intrinsic relatedness, so should this be reflected by having similar concepts closer.
The need not necessarily lie close along all dimensions, but similar concepts should be consistently located closely in some subset of dimensions.
This aids direct data mining from the VS representation, and is a good indicator of the representation's success.
Note that different kinds of similarities may be found along different subsets of dimensions; there may be useful information in the dimensions over which certain interactions are placed.

\section{Absolute}
Runs over similar datasets place most points in similar absolute locations.
For example, a 10-dimensional vector for the word ``confused" based on the Brown corpus should be contain roughly the same values even if a few extra words are tagged on to the end of the data.
This assumes that the data generation is already deterministic.
While affording the concept that languages and text types converge on certain points given enough data, this requirement also enables many comparative studies; for example, longitudinal studies become possible, allowing sense differences to be detected as the observation (i.e. the usage of words in text, through corpora) shifts over time.

\section{Semantic dimensions}
The first question asked by many computer scientists outside of ML/NLP, when hearing of embeddings and their power, is simple: what do these dimensions mean?
It is not a question that is, as yet, well-addressed.
This paper suggests that dimensions should have some meaning, and indeed, recent VS representation evaluation tools (e.g.~\newcite{tsvetkov2015evaluation}) have raised and focused on achieving exactly that.
For example, on dimension may reflect spatial vs. temporal meanings, another singular vs. plural, and another go from countries down to villages.
Of course it is unlikely that semantic vectors between individual word representation will traverse only one or two dimensions, but there is value in ensuring that the most part of a difference corresponding to a dimension's meaning is concentrated in that dimension.
Fixing word representations in absolute, stable, repeatable geometries, and then being able to analyse and identify the roles of individual dimensions, again enables a much broader range of analytical methods to operate over them than currently possible.

\section{Similar scales}
Dimensions should have similar scales.
Rotating the dataset in its Hibbert space should retain the geometry; vector distances between concepts should still permit e.g. the analogy results.
This only works if all dimensions are operating along similar scales, which is a critical requirement if the VS representation needs to be rotated or otherwise preprocessed to concentrate semantic scales in each their own dimension.

\section{Fast}
It should be as fast as possible to generate the representation, and implementations should use available modern hardware effectively.
For example, an algorithm that can take effective advantage of multi-threading, NUMA and caches, GPGPU power, or even offer the option of terminating earlier or later with a corresponding difference in quality~\cite{stratos2014spectral}, is preferable to one that doesn't.

\section{Streaming support}
Ideally, the method for generating a VS representation should support streaming operations.
This means that deletion and insertion of observations needs to be possible.
Language changes over time and concept drift renders representations out of date; the faster-moving the text type, the quicker this happens.
For example, models trained on 2012 tweets for NER perform progressively worse each year after 2012.
To support streaming, operations of text addition and deletion need to be supported in constant time.
This way, content can be added and word positions recalculated as new text becomes available; and to maintain a fresh model, it has to be possible to remove older text.

\section{Conclusion}

This proposal identified desiderata for future vector-space representations of words.
Many of these are perhaps challenging, though implementing them opens up the world of embeddings to the wider field of analysis outside of the NLP/ML community, while also enabling powerful new modes of generating and reasoning about words.

\bibliography{desiderata}

\begin{thebibliography}{}

\bibitem[\protect\citename{Derczynski and
  Chester}2016]{derczynski2016generalised}
Leon Derczynski and Sean Chester.
\newblock 2016.
\newblock Generalised brown clustering and roll-up feature generation.
\newblock In {\em Proceedings of the conference of the Association for
  Advancement of Artificial Intelligence}. AAAI.

\bibitem[\protect\citename{Mikolov \bgroup et al.\egroup
  }2013]{mikolov2013distributed}
Tomas Mikolov, Ilya Sutskever, Kai Chen, Greg~S Corrado, and Jeff Dean.
\newblock 2013.
\newblock Distributed representations of words and phrases and their
  compositionality.
\newblock In {\em Advances in neural information processing systems}, pages
  3111--3119.

\bibitem[\protect\citename{Schnabel \bgroup et al.\egroup
  }2015]{schnabel2015evaluation}
Tobias Schnabel, Igor Labutov, David Mimno, and Thorsten Joachims.
\newblock 2015.
\newblock Evaluation methods for unsupervised word embeddings.
\newblock In {\em Proc. of EMNLP}. ACL.

\bibitem[\protect\citename{Stratos \bgroup et al.\egroup
  }2014]{stratos2014spectral}
Karl Stratos, Do-kyum Kim, Michael Collins, and Daniel Hsu.
\newblock 2014.
\newblock A spectral algorithm for learning class-based n-gram models of
  natural language.
\newblock In {\em Proceedings of the Association for Uncertainty in Artificial
  Intelligence}.

\bibitem[\protect\citename{Tsvetkov \bgroup et al.\egroup
  }2015]{tsvetkov2015evaluation}
Yulia Tsvetkov, Manaal Faruqui, Wang Ling, Guillaume Lample, and Chris Dyer.
\newblock 2015.
\newblock Evaluation of word vector representations by subspace alignment.
\newblock In {\em Proc. of EMNLP}. ACL.

\end{thebibliography}
\bibliographystyle{acl2016}

\end{document}